\documentclass[10pt,twocolumn,letterpaper]{article}

\usepackage[final]{cvpr}      
\usepackage{amsmath}
\usepackage{amssymb}
\usepackage{booktabs}
\usepackage{url}

\usepackage{times,graphicx,amsmath,amssymb,booktabs,tabulary,multirow,overpic,xcolor}
\usepackage{colortbl}
\usepackage{makecell}
\usepackage{float}
\usepackage{bbding}

\definecolor{sh_gray}{rgb}{0.84,0.84,0.84}
\definecolor{sh_gray2}{rgb}{1,0.89,0.75}
\definecolor{color3}{rgb}{0.95,0.95,0.95}
\definecolor{color4}{rgb}{0.96,0.96,0.86}
\definecolor{color5}{rgb}{0.90,0.90,0.90}
\usepackage[pagebackref,breaklinks,colorlinks]{hyperref}
\usepackage[capitalize]{cleveref}
\usepackage{multirow}
\crefname{section}{Sec.}{Secs.}
\Crefname{section}{Section}{Sections}
\Crefname{table}{Table}{Tables}
\crefname{table}{Tab.}{Tabs.}
\usepackage{algorithm}
\usepackage{algorithmic}

\begin{document}

\title{AI-driven Prediction of Insulin Resistance in Normal Populations: Comparing Models and Criteria}

\author{%
	Weihao Gao$^{1}$ , Zhuo Deng$^{1}$, Zheng Gong$^{1}$, Ziyi Jiang$^{1}$, Lan Ma$^{1,}$\thanks{$ $ = corresponding author \\Email: malan@sz.tsinghua.edu.cn}~\\ 
    $^{1}$Shenzhen International Graduate School, Tsinghua University\\
    }

\maketitle

\begin{abstract}

\textcolor{purple}{Background}

Insulin resistance is a critical precursor to diabetes and significantly increases the risk of cardiovascular disease. Traditional methods for insulin assessing resistance often require multiple blood tests, including fasting blood glucose. Developing a simple AI-based model that relies on minimal invasive tests could greatly facilitate health monitoring in non-diabetic populations.

\textcolor{purple}{Methods}

This study aims to develop an AI-driven model that uses only fasting blood glucose as an invasive measure to predict insulin resistance in non-diabetic populations. We constructed four AI-based prediction tasks using data from the National Health and Nutrition Examination Survey (NHANES) database (9199-2020). These tasks were based on three commonly used insulin resistance (IR) indicators: HOMA-IR, TyG, and METS-IR. To test the generalizability of our models, we performed cross-dataset validation using data from the China Health and Retirement Longitudinal Study (CHARLS, 2015). We selected simple and accessible input features, including age, gender, height, weight, pulse, blood pressure, waist circumference, and fasting blood glucose, to build our prediction models. Additionally, we used SHAP values to interpret the contributions of these features to the predictions.

\textcolor{purple}{Results}

In our study, the CatBoost algorithm achieved excellent performance in classification tasks. For numerical prediction of the METS-IR index, neural networks, particularly TabKANet, demonstrated superior performance in cross-dataset validation. In the NHANES test set, the AUC values for predicting insulin resistance were 0.8596 (HOMA-IR index) and 0.7777 (TyG index), with an external validation AUC of 0.7442 for the TyG index. For METS-IR prediction, our model achieved AUC values of 0.9731 (internal) and 0.9591 (external). Additionally, the AI-driven model for predicting METS-IR had RMSE values of 3.2643 (internal) and 3.057 (external). SHAP analysis revealed that waist circumference was a key factor in predicting insulin resistance, highlighting its importance in early diabetes and cardiovascular disease prediction.

\end{abstract}

\section*{Background}

Diabetes is one of the leading diseases that cause death and disability worldwide, affecting people of all ages and sexes around the world. With the increase in the global obesity rate and the aggravation of population aging, it is predicted that the prevalence of diabetes will increase significantly by 2050. More than 1.3 billion people are expected to have diabetes, and the age-standardized total diabetes prevalence rate will exceed 10\%\cite{zhou2024burden,ong2023global}. Thus, diabetes poses a significant public health challenge globally, necessitating effective preventive and management strategies.

Insulin resistance (IR) is an important biomarker for recognizing the early stages of diabetes and also a central pathophysiological mechanism underlying the development of type 2 diabetes mellitus. It refers to the reduced efficiency of insulin in promoting glucose absorption and utilization by skeletal muscle, liver, and adipose tissue cells\cite{hou2024association,tsai2023development}. This occurs when insulin levels are insufficient to meet the body's glucose demands. Consequently, pancreatic beta cells must compensate by secreting more insulin to maintain normal blood glucose levels. As insulin resistance intensifies, the number and function of beta cells gradually decrease, leading to an increase in blood sugar levels. The occurrence and development of insulin resistance may be related to multiple factors, including obesity, stress, certain medications (such as steroids), pregnancy, insulin antibodies, and genetic variations in the insulin signaling pathway.

Studies have shown that the decline in beta cell function begins approximately 12 years before the diagnosis of diabetes mellitus (DM) and continues to deteriorate throughout the disease\cite{colagiuri2002lower,simonson2011international}. However, the progression of insulin resistance (IR) exists before the decline of beta cell function. Therefore, IR is the earliest warning signal for DM. Early identification and warning of IR in the population can help reduce the risk of diabetes or delay the time of diagnosis for people of all age groups. Early intervention for IR or DM can also significantly improve the prognosis of patients with diabetes\cite{uk1998intensive,advance2008intensive,duckworth2009glucose,action2008effects}.

Beyond diabetes, insulin resistance has been recognized as a standalone risk factor for the development of cardiovascular diseases\cite{ausk2010insulin,moshkovits2021association,shinohara2002insulin,mohan2014insulin,tan2023association}. Given the high prevalence and mortality of cardiovascular diseases, IR represents a significant public health burden, further highlighting the importance of early detection and intervention.

Therefore, from the perspective of prevention and management of diabetes and reduction of cardiovascular disease mortality, it is very important to detect and warn the general population of insulin resistance. Although blood glucose clamp technology is considered the gold standard for evaluating IR, due to its high analysis cost and complex operating procedures, this technology is mainly limited to small-scale studies and has not been widely applied in large-scale epidemiological investigations\cite{minh2021assessment}. To address this issue, researchers have proposed various alternative indicators. At present, commonly used insulin resistance replacement indices in clinical practice include: the homeostasis model assessment of insulin resistance (HOMA-IR), triglyceride glucose index (TyG index), and insulin resistance metabolic score (METS-IR)\cite{duan2024metabolic,bello2018mets,tahapary2022challenges,tao2022triglyceride}.

HOMA-IR is one of the most widely used alternative indicators of insulin resistance, which is calculated by combining fasting blood glucose and fasting plasma insulin. However, the evaluation effectiveness of this model varies by population race and shows some limitations in patients receiving insulin therapy or those with beta cell dysfunction\cite{tao2022triglyceride,hou2024association}. In addition, calculating HOMA-IR relies on laboratory measurements of fasting insulin levels, which is difficult to achieve in resource limited countries and limits its widespread use in daily clinical practice.

The TyG index is calculated based on triglycerides and fasting blood glucose, which is relatively simple and easy to obtain, and is considered a time-saving and relatively simple IR marker. Multiple studies have shown that it is consistent or better than HOMA-IR in evaluating IR\cite{chamroonkiadtikun2020triglyceride,simental2008product}. Therefore, extensive research has been conducted to explore the relationship between TyG index and cardiovascular disease and its prognosis. Some research reports indicate that the TyG index is significantly correlated with overall mortality and cardiovascular mortality in the general population, especially in the population under 65 years old\cite{chen2023association}. However, some studies have reported inconsistent results, stating that there is no significant relationship between TyG index and all-cause or cardiovascular mortality\cite{liu2022relationship,duan2024metabolic}.

METS-IR, as another relatively novel IR assessment standard, combines fasting blood glucose, BMI, triglycerides, and high-density lipoprotein cholesterol\cite{bello2018mets}. In recent studies, it has been reported that METS-IR exhibits better predictive ability for all-cause mortality and cardiovascular mortality compared to HOMA-IR and TyG index. Specifically, when the baseline value of METS-IR is below 41.33, it is negatively correlated with the mortality rate; When the baseline value of METS-IR is higher than 41.33, it is positively correlated with mortality. This association is particularly significant in the non-elderly population under 65\cite{duan2024metabolic}.

IR evaluation is of great importance, including early warning of the progress of diabetes and cardiovascular disease. However, in clinical practice, different indicators are not applied enough, especially for adults who are not diagnosed with diabetes. The above three criteria for assessing insulin resistance require at least two laboratory test results based on blood samples, which especially restricts scheduled health monitoring and evaluation of the group with unconfirmed diabetes. 

In recent years, artificial intelligence has been increasingly employed in the management of diabetes, including the detection and management of diabetic retinopathy, as well as AI - driven prediction of diabetes\cite{howlader2022machine,joshi2021predicting}. Despite these advances, studies focused on predicting IR, particularly in undiagnosed diabetic populations, remain limited. In the past, researchers attempted to predict HOMA-IR using information such as BMI, fasting blood glucose, triglycerides, and high-density lipoprotein cholesterol, achieving a performance of AUC 0.87\cite{lee2022development,park2022development}. Nevertheless, such work still relies on multiple blood-based laboratory tests. Although it can serve as an alternative to hospital laboratory tests, it cannot achieve home self-assessment of IR for the widest population.

Given these challenges and the potential of AI in healthcare, our study aims to develop a novel, accessible, and accurate method for IR prediction using minimally invasive tests and home-measurable features. To meet the needs of large-scale home self-testing, we set strict conditions, using only fasting blood glucose as the input indicator. We developed and tested the system using data from the National Health and Nutrition Examination Survey (NHANES) and validated it in the China Health and Nutrition Survey (CHALES) database and nationwide population. We systematically assessed the predictive performance of AI-based methods in a non-diabetic population under strict conditions, using various insulin resistance criteria, including HOMA-IR, the TyG index, and METS-IR. The experimental results demonstrate that combining fasting blood glucose, an invasive test, with physical characteristics that can be measured at home, achieves accurate assessment under different diagnostic criteria for IR. In particular, the prediction of insulin resistance associated with METS-IR shows remarkable performance. Using AI methods, we achieved a classification diagnosis with an area under the curve (AUC) exceeding 0.97. Additionally, we were able to make numerical predictions of METS-IR, with an internal validation R2 of 0.916 and an external test set R2 of 0.8183.

\section*{Methods}

\subsection*{Study design and participants}


In this research, we utilized data from two prominent health surveys: the United States' National Health and Nutrition Examination Survey (NHANES) and China's Health and Retirement Longitudinal Study (CHARLS).

NHANES is a comprehensive health survey conducted regularly by the Centers for Disease Control and Prevention (CDC) and the National Center for Health Statistics (NCHS). The data are freely available to the public and widely used in epidemiological studies, health policy evaluations, and the development of public health interventions. The NHANES research protocol has been approved by the NCHS Institutional Review Board, and all participants or their representatives have provided written informed consent. The survey includes anthropometric measurements, health and nutrition questionnaires, and laboratory tests, with participants completing the questionnaires during home interviews. For this study, we included NHANES participants from January 1, 1999, to March 31, 2020. Participants under 18 years old, those with incomplete laboratory data, or those with diabetes were excluded from the analysis. This resulted in a final sample of 22,008 participants, with relevant information including age, gender, race, height, weight, BMI, waist circumference, pulse, blood pressure, fasting glucose, fasting serum insulin, high-density lipoprotein cholesterol, and triglycerides. The dataset was divided into training, validation, and test sets in a 6:2:2 ratio.

We used the NHANES dataset to construct AI models for different insulin resistance assessment criteria and validated them using cross-dataset, cross-country data from CHARLS. CHARLS is a nationwide population-based cohort study targeting Chinese adults aged 45 and above\cite{chen2019venous}. CHARLS conducted five regular surveys between 2011 and 2020. Participants were recruited from both rural and urban areas through a multi-stage stratified probability proportional to size sampling strategy, covering 150 counties or districts in 28 provinces across China. The CHARLS study adheres to the principles of the Declaration of Helsinki and has been approved by the Institutional Review Board of Peking University (IRB00001052-11015). All participants provided written informed consent before participating in the CHARLS study. Standardized questionnaires were used to collect information on socio-demographic characteristics, medical history, health behaviors, cognitive function, and depressive status through face-to-face interviews. We analyzed participants from the 2015 survey, excluding those with diabetes and incomplete data, resulting in a final sample of 10,333 participants with information on age, gender, height, weight, BMI, pulse, blood pressure, fasting glucose, high-density lipoprotein cholesterol, and triglycerides. This dataset served as the external validation set for this study.

\begin{figure*}[h]
\begin{center}
    \includegraphics[width=\textwidth]{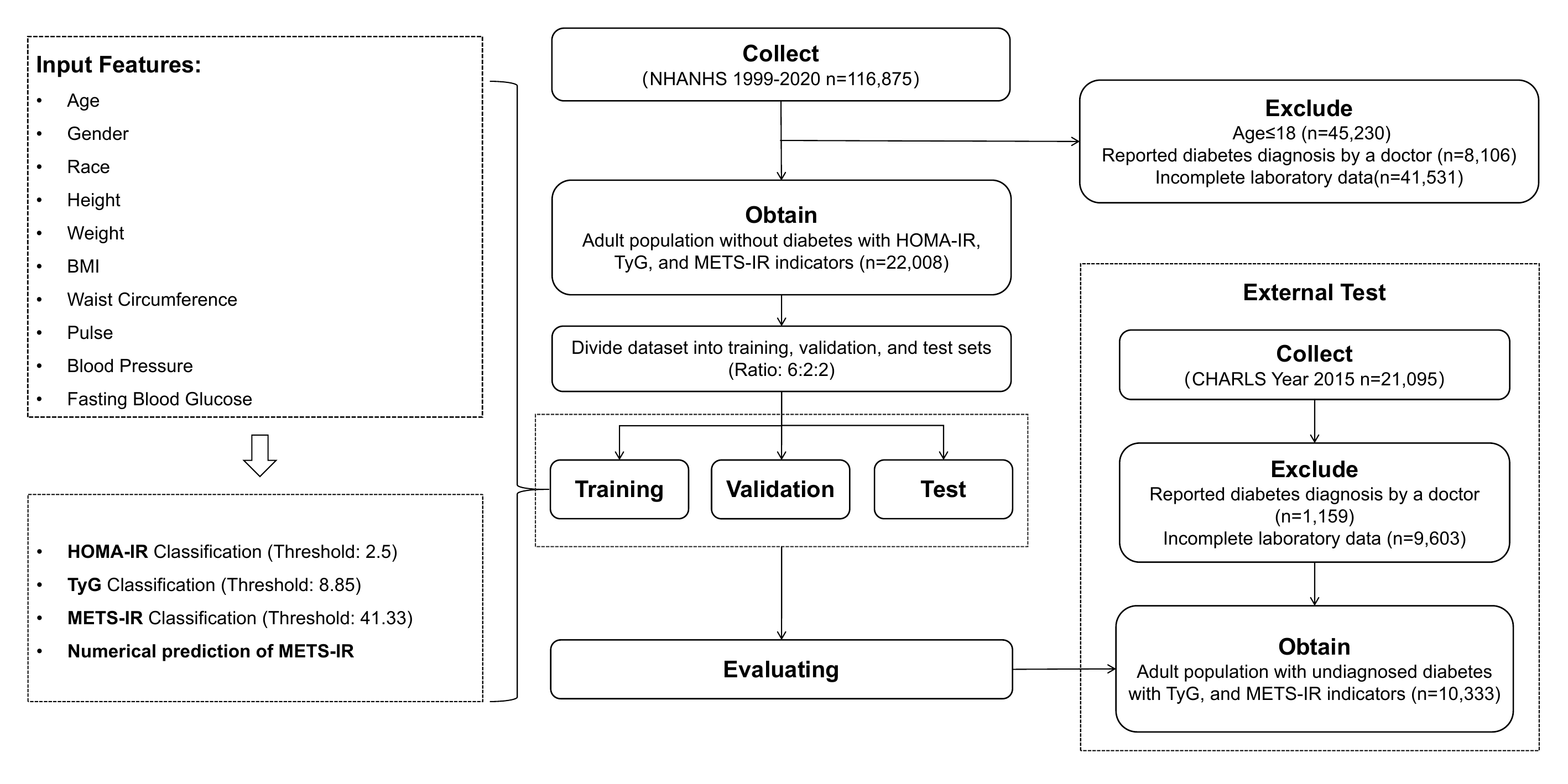} 
\end{center}
\caption{Flow diagram of study population selection process.}
\label{fig:flow}
\end{figure*}

The final study population consisted of 32,341 participants. The patient selection process is shown in Fig. \ref{fig:flow}.

\subsection*{Different IR assessment replacement index}

To explore the potential of artificial intelligence methods to assess insulin resistance relying solely on fasting blood glucose as the single invasive test, three different IR assessment replacement indexes were included in this study, namely the HOMA-IR, the TyG index, and the METS-IR index.




The formula for HOMA-IR is given by:
\[ \text{HOMA-IR} = \frac{\text{FPG} \times \text{FI}}{22.5} \]
where FPG = Fasting Plasma Glucose (mmol/L), FI = Fasting Insulin (mIU/L).

The TyG index can be calculated using the following formula:
\[ \text{TyG Index} = \ln \left( \frac{\text{TG} \times \text{FPG}}{2} \right) \]
where TG = Triglycerides (mg/dL), FPG = Fasting Plasma Glucose (mg/dL).

The METS-IR is calculated using the following formula:
\[ \text{METS-IR} = \frac{\ln \left( 2 \times \text{FPG} + \text{TG} \right) \times \text{BMI}}{\ln (\text{HDL-C})} \]
where FPG = Fasting Plasma Glucose (mg/dL), TG = Triglycerides (mg/dL), BMI = Body Mass Index, HDL-C = High-Density Lipoprotein Cholesterol (mg/dL).

Based on previous studies of various insulin resistance evaluation indicators, the following cut-off values have been established: For the HOMA-IR indicator, a threshold of 2.5 is commonly used to define IR in adults\cite{muniyappa2008current}. For the TyG indicator, an optimal cut-off value of 8.85 has been identified for predicting metabolic syndrome\cite{kim2021triglyceride}. For the METS-IR indicator, a critical value of 41.33 has been proposed\cite{duan2024metabolic}. Specifically, a baseline METS-IR value below 41.33 is negatively correlated with mortality, whereas a value above this threshold is positively correlated with mortality.


This study evaluates IR based on simplified input characteristics from the non-diabetic population. We constructed four different tasks aimed at assessing various IR indicators. The specific tasks are as follows:

\begin{itemize}
    \item \textbf{IR assessment based on HOMA-IR}: A binary classification assessment with a threshold of 2.5.
    \item \textbf{IR assessment based on TyG}: A binary classification assessment with a threshold of 8.85.
    \item \textbf{IR assessment based on METS-IR}: A binary classification assessment with a threshold of 41.33.
    \item \textbf{Numerical prediction of METS-IR}: This is a numerical regression task that aims to predict the specific value of the METS-IR index.
\end{itemize}

The first three tasks are independent binary classification tasks, while the fourth task is a numerical regression task. This study explores the potential of different AI methods to assess insulin resistance in a normal population. Specifically, we aim to evaluate the performance of AI models constructed with minimal input features to meet the needs of the broadest population for frequent self-assessment. Additionally, we aim to assess the accuracy and generalizability of these models across different datasets, particularly those from different countries.


\subsection*{AI model selection}

To construct an AI-based decision model for IR in nondiabetic individuals, we selected a range of traditional machine learning models, including Logistic Regression\cite{lavalley2008logistic}, Random Forest\cite{breiman2001random}, XGBoost\cite{chen2016xgboost}, and CatBoost\cite{prokhorenkova2018catboost}, as well as neural network models such as MLP\cite{hornik1989multilayer}, TabTransformer\cite{huang2020tabtransformer}, and TabKANet\cite{gao2024tabkanet}. For the numerical prediction task of METS-IR, we used Linear Regression\cite{montgomery2021introduction} instead of Logistic Regression. This choice is based on the fact that linear regression is well-suited for modeling continuous numerical outputs by minimizing the mean squared error, whereas logistic regression is primarily designed for classification tasks and cannot directly address continuous numerical prediction.

Our input features consist of nine dimensions, with race and gender as categorical features, and the remaining features as continuous numerical variables. This is a typical prediction task based on tabular data, which has traditionally relied on classical decision tree models. In the field of tabular data modeling, tree-based models such as XGBoost\cite{chen2016xgboost} and CatBoost\cite{prokhorenkova2018catboost} have long been the leading methods, outperforming traditional machine learning approaches like logistic regression or random forests. Despite recent advancements in neural network architectures, tree ensemble-based methods are still widely regarded as the state-of-the-art for tasks such as classification or regression\cite{chen2016xgboost, prokhorenkova2018catboost}. This perception stems from the observation that tree-based ensemble models typically offer competitive predictive accuracy and faster training speeds.

The Multilayer Perceptron (MLP) has long been a core component of neural networks. With its fully connected architecture, MLPs can effectively approximate complex functions and have been widely favored in numerous applications due to their strong expressive power. However, the MLP architecture also has certain limitations. For example, its activation functions are typically fixed, which may limit the model's flexibility in capturing complex relationships within the data, as it relies on predefined nonlinear functions. TabTransformer is a novel model that embeds categorical features from the input data and further learns using the Transformer architecture\cite{huang2020tabtransformer}. TabKANet, on the other hand, is a more advanced neural network-based model that integrates numerical features using a Kolmogorov-Arnold Network, combines them with categorical embedding structures, and learns through the Transformer architecture\cite{liu2024kan}. Both TabTransformer and TabKANet leverage the Transformer for feature learning, making them computationally more complex but also endowing them with greater capability for handling complex data relationships.

In this study, we set the number of trees for both XGBoost and CatBoost models to 1,000, with a maximum tree depth of 8. For the neural network models, the embedding dimension for both TabTransformer and TabKANet was set to 64. During the training process, TabTransformer and TabKANet employed the Transformer module with multi-head attention set to 8 and the number of layers set to 3. For the neural networks, the classification tasks used CrossEntropyLoss as the loss function, with a learning rate of 1e-3 and the AdamW optimizer for parameter updates. The regression tasks used MSELoss as the loss function, with a learning rate of 1e-4 and the SGD optimizer for training.

\section*{Results}

\subsection*{Study data characteristics and data split}

This study was based on the NHANES dataset, which collected data from January 1999 to March 2020 (prior to the pandemic). We excluded individuals under 18 years old, those diagnosed with diabetes, and those with missing data. The final NHANES dataset included 22,008 participants, which were randomly divided into training, validation, and testing sets in a 6:2:2 ratio. In addition to the internal testing dataset from NHANES, we introduced data from the 2015 China Health and Retirement Longitudinal Study (CHARLS) as an external validation dataset for cross-national and cross-dataset verification. Table \ref{table:alldata} summarizes the characteristics of the development and external testing datasets, while Figure \ref{fig:flow} illustrates the overall workflow of this study.

Due to the absence of plasma insulin level data in the CHARLS dataset, external evaluation of HOMA-IR could not be conducted in subsequent experiments. The development dataset (NHANES) and the external validation dataset (CHARLS) were organized by different teams and based on populations from the United States and China, respectively. NHANES is a multi-ethnic dataset, whereas all samples in the CHARLS dataset were categorized under the "Other Races (including multi-ethnic)" category. Additionally, while the target population of CHARLS is Chinese adults aged 45 and above, the minimum age in the NHANES dataset is 19 years old. These differences highlight the significant variability between the two datasets in terms of demographic characteristics.


\begin{table*}[h!]
\centering
\caption{Characteristics of the participants}
    \resizebox{\textwidth}{!}{

\small
\begin{tabular}{llllll}
\hline
Characteristics & NHANES & Train & Val & Test & CHARLS\\ \hline

Number of participants & 22,008 & 13,205 & 4,402 & 4,401 & 10,333 \\
Age (years) & 46.82$\pm$18.28 & 46.75$\pm$18.33 & 46.82$\pm$18.29 & 46.99$\pm$18.12 & 60.20$\pm$9.87 \\
Female & 1,1433 (51.94\%) & 6,826 (51.69\%) & 2,321 (52.72\%) & 2,286 (51.04\%) & 5,544 (53.65\%) \\
Body mass index (kg/m$^2$) & 28.41$\pm$6.53 & 28.41$\pm$6.59 & 28.48$\pm$6.44 & 28.34$\pm$6.43 & 23.78$\pm$3.84 \\
Fasting plasma glucose (mg/dl) & 100.64$\pm$18.98 & 100.61$\pm$19.17 & 100.67$\pm$19.50 & 100.68$\pm$17.85 & 93.97$\pm$11.76 \\
Plasma insulin level (uU/mL) & 12.26$\pm$11.32 & 12.21$\pm$11.48 & 12.34$\pm$10.97 & 12.31$\pm$11.17 & - \\
Triglycerides (mg/dl) & 117.01$\pm$66.69 & 116.84$\pm$66.79 & 117.76$\pm$66.99 & 116.74$\pm$66.11 & 136.27$\pm$85.59 \\
HDL Cholesterol (mg/dl) & 54.58$\pm$15.97 & 54.64$\pm$16.11 & 54.47$\pm$15.97 & 54.51$\pm$15.53 & 51.66$\pm$11.37 \\
Systolic (mm Hg) & 122.44$\pm$18.49 & 122.43$\pm$18.60 & 122.48$\pm$18.05 & 122.41$\pm$18.63 & 127.27$\pm$19.33 \\
Diastolic (mm Hg) & 70.45$\pm$11.77 & 70.31$\pm$11.88 & 70.83$\pm$11.50 & 70.48$\pm$11.71 & 75.35$\pm$11.22 \\
Pulse (60 sec.) & 70.66$\pm$11.74 & 70.64$\pm$11.77 & 70.65$\pm$11.70 & 70.76$\pm$11.67 & 73.58$\pm$10.45 \\
HOMA-IR & 3.17$\pm$3.60 & 3.16$\pm$3.68 & 3.19$\pm$3.51 & 3.20$\pm$3.45 & - \\
TyG & 8.52$\pm$0.59 & 8.51$\pm$0.59 & 8.52$\pm$0.58 & 8.52$\pm$0.58 & 8.60$\pm$0.56 \\
METS-IR & 41.68$\pm$11.49 & 41.67$\pm$11.57 & 41.84$\pm$11.47 & 41.55$\pm$11.26 & 35.14$\pm$7.17 \\
Mexican American & 3,805(17.28\%) & 2,293(17.36\%) & 738(16.76\%) & 773(17.56\%) & - \\
Other Hispanic & 1,868(8.48\%) & 1,107(8.38\%) & 377(8.56\%) & 384(8.72\%) & - \\
Non-Hispanic White & 9,688(44.02\%) & 5,784(43.80\%) & 2,001(45.45\%) & 1,903(43.24\%) & - \\
Non-Hispanic Black & 4,401(19.99\%) & 2,687(20.34\%) & 832(18.90\%) & 882(20.04\%) & - \\
Other Race-Including Multi-Racial & 2,247(10.21\%) & 1,334(10.10\%) & 454(10.31\%) & 459(10.42\%) & 1,0333(100\%) \\

\end{tabular}
}
\label{table:alldata}
\end{table*}

\subsection*{Comparative Performance in HOMA-IR classification}

\begin{figure*}[h]
\begin{center}
    \includegraphics[width=\textwidth]{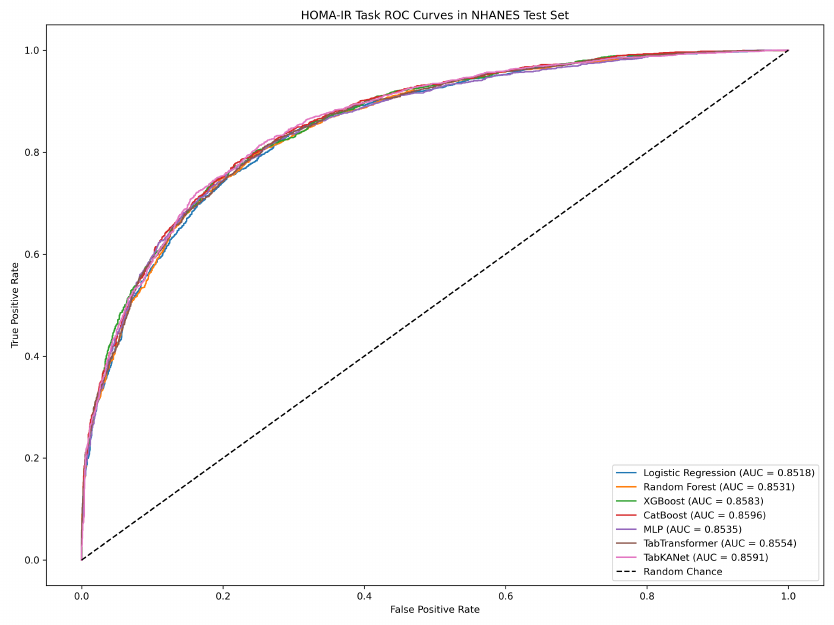} 
\end{center}
\caption{ROC curves of HOMA-IR classification using different methods. Due to the lack of Plasma insulin levels in the CHARLS dataset, external validation calculations cannot be performed.}
\label{fig:homa}
\end{figure*}

Table \ref{table:homatask} presents the performance of various AI methods in predicting the classification of HOMA-IR among normal individuals. The characteristics of participants in the data split for HOMA-IR are shown in Supplementary Table 1. Among the different methods, the CatBoost algorithm, which is based on Gradient Boosting Decision Trees (GBDT), achieved the best performance with an Area Under the Curve (AUC) value of 0.8583. Meanwhile, the TabKANet method, which is based on neural networks, also achieved performance very close to that of CatBoost, with an AUC value of 0.8691. Overall, in the internal testing of the NHANES database, various methods were able to effectively differentiate the state of insulin resistance assessed by HOMA-IR. Whether it was traditional machine learning methods, decision tree models, or more computationally complex neural network methods, all models achieved AUC values above 0.85, demonstrating good predictive capabilities. Figure \ref{fig:homa} further displays the Receiver Operating Characteristic (ROC) curves of the different models, intuitively showing the performance differences among them.


\begin{table*}[h!]
	\footnotesize
	\centering	
    \caption{The performance of different model methods in HOMA-IR task.}
	\scalebox{1}{
			\begin{tabular}{c|c c c c c}
				\toprule
		Model & AUC & ACC & F1 & Precision & Recall \\
				\midrule  
Logistic Regression & 0.8518 & 0.7741 & 0.7689 & 0.7741 & 0.7668 \\
Random Forest & 0.8531 & 0.7768 & 0.7728 & 0.7753 & 0.7715 \\
XGBoost & 0.8583 & 0.7802 & 0.7768 & 0.7783 & 0.7758 \\
CatBoost & 0.8596 & 0.7802 & 0.7773 & 0.7779 & 0.7768 \\
MLP & 0.8535 & 0.7766 & 0.7721 & 0.7710 & 0.7109 \\
TabTransformer & 0.8554 & 0.7760 & 0.7689 & 0.7940 & 0.6728 \\
TabKANet & 0.8591 & 0.7769 & 0.7755 & 0.7348 & 0.7827 \\ 
                \bottomrule
	    \end{tabular}}
	\label{table:homatask}
\end{table*}

\subsection*{Comparative Performance in TyG Index Classification}

Table \ref{table:tygtask} presents the performance of various methods in predicting the classification of the TyG index among normal individuals, with the threshold for the TyG index set at 8.85. The characteristics of participants in the data split for TyG are shown in Supplementary Table 2. Among these methods, the CatBoost algorithm achieved the best performance. In the internal test set, CatBoost achieved an AUC value of 0.7777, while in the CHARLS external test set, its AUC value was 0.7442. Meanwhile, the TabKANet method also achieved performance very close to that of CatBoost. In the internal test set, the AUC value of TabKANet was 0.7696, and in the external test set, its AUC value was 0.7441.

\begin{figure*}[h]
\begin{center}
    \includegraphics[width=\textwidth]{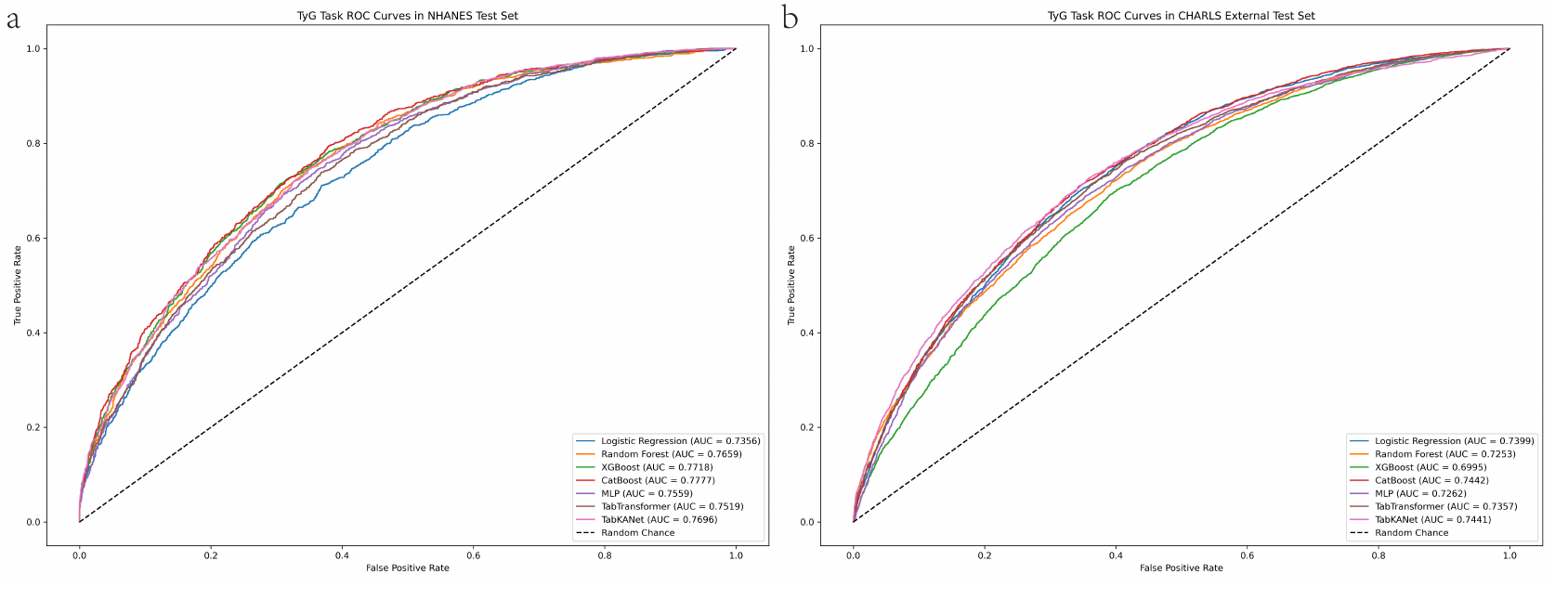} 
\end{center}
\caption{ROC curves of TyG classification using different methods. a) NHANES test set results. b) CHARLS external test results.}
\label{fig:tyg}
\end{figure*}

Overall, even though CHARLS is a completely cross-national and cross-dataset external validation experiment, the AUC values of the models only showed a slight decrease between internal and external tests. In particular, the performance of the outstanding AI methods, such as CatBoost and TabKANet, remained stable in the cross-dataset tests. Figure \ref{fig:tyg} further displays the ROC curves of the different models in the NHANES internal test set and the CHARLS dataset, intuitively showing the performance differences of the models across different datasets.


\begin{table*}[h!]
	\footnotesize
	\centering	
    \caption{Comparative Performance in TyG classification}
    \resizebox{\textwidth}{!}{
			\begin{tabular}{c|ccccc|ccccc}
				\toprule
                
				\multirow{2}{*}{Model}  &\multicolumn{5}{c|}{Internal Test} &\multicolumn{5}{c}{CHARLS(External Test)} \\
				& AUC & ACC & F1 & Precision & Recall&AUC & ACC & F1 & Precision & Recall \\
				\midrule  
Logistic Regression & 0.7356 & 0.7375 & 0.6030 & 0.6782 & 0.5983 & 0.7399 & 0.6991 & 0.6563 & 0.6585 & 0.6545\\
Random Forest & 0.7659 & 0.7502 & 0.6151 & 0.7111 & 0.6084
 & 0.7253 & 0.6907 & 0.6476 & 0.6492 & 0.6462 \\
XGBoost & 0.7718 & 0.7496 & 0.6547 & 0.6920 & 0.6429
& 0.6995 & 0.6640 & 0.6260 & 0.6246 & 0.6279\\
CatBoost & 0.7777 & 0.7564 & 0.6568 & 0.7056 & 0.6437 
& 0.7442 & 0.7064 & 0.6603 & 0.6589 & 0.6620\\
MLP & 0.7559 & 0.7426 & 0.6447 & 0.5813 & 0.3781 
& 0.7262 & 0.7009 & 0.6509 & 0.6507 & 0.6510 \\
TabTransformer & 0.7519 & 0.7464 & 0.6405 & 0.6013 & 0.3536
 & 0.7357 & 0.6927 & 0.6576 & 0.6536 & 0.6674\\
TabKANet & 0.7696 & 0.7510 & 0.6276 & 0.6423 & 0.3047
& 0.7441 & 0.7152 & 0.6645 & 0.6601 & 0.6632\\
                \bottomrule
	    \end{tabular}}
	\label{table:tygtask}
\end{table*}

\subsection*{Comparative Performance in METS-IR classification}

Table \ref{table:loosemetstask} presents the performance of various artificial intelligence methods in predicting the METS-IR classification among normal individuals. The characteristics of participants included in the data split for METS-IR are detailed in Supplementary Table 3.

\begin{figure*}[h]
\begin{center}
    \includegraphics[width=\textwidth]{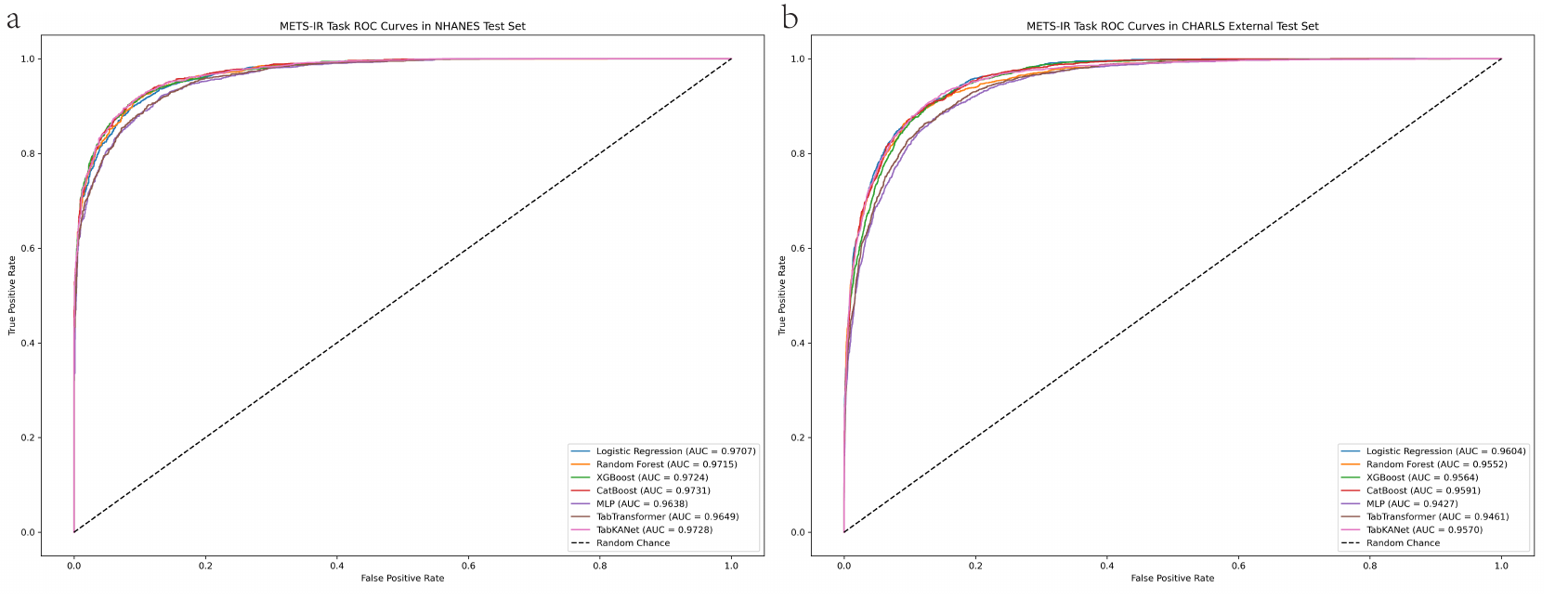} 
\end{center}
\caption{ROC curves of METS-IR classification using different methods. a) NHANES test set results. b) CHARLS external test results.}
\label{fig:metsloose}
\end{figure*}

All methods achieved good performance. Among them, CatBoost achieved an AUC value of 0.9731 in the internal test and 0.9501 in the CHARLS external validation. Figure \ref{fig:metsloose} further displays the ROC curves of the models in the NHANES internal test set and the CHARLS dataset.

To further evaluate the generalizability of our model across diverse populations, we conducted subgroup analyses based on race/ethnicity using the NHANES dataset. The model demonstrated robust performance across all subgroups, with AUC values ranging from 0.96 to 0.97 for METS-IR classification (Supplementary Figure 1). Notably, the model maintained high accuracy in minority populations, such as non-Hispanic Black and Hispanic individuals, who are often underrepresented in insulin resistance studies. These results suggest that our AI-driven approach applies to a wide range of ethnic groups, supporting its potential for global health applications.

Overall, different AI methods were all able to complete the insulin resistance assessment based on METS-IR in the general population with high performance. This assessment relies solely on fasting blood glucose as the only invasive test indicator. Although neural network methods are more complex in model construction and consume more computational resources in training and inference, they did not show significant advantages in the classification problem.




\begin{table*}[h!]
	\footnotesize
	\centering	
    \caption{Comparative Performance in METS-IR classification}
    \resizebox{\textwidth}{!}{
			\begin{tabular}{c|ccccc|ccccc}
				\toprule
                
				\multirow{2}{*}{Model}  &\multicolumn{5}{c|}{Internal Test} &\multicolumn{5}{c}{CHARLS(External Test)} \\
				& AUC & ACC & F1 & Precision & Recall&AUC & ACC & F1 & Precision & Recall \\
				\midrule  

Logistic Regression & 0.9707 & 0.9041 & 0.9030 & 0.9039 & 0.9023& 0.9604 & 0.9188 & 0.8508 & 0.8825 & 0.8267 \\
Random Forest & 0.9715 & 0.9059 & 0.9048 & 0.9056 & 0.9042 & 0.9552 & 0.9167 & 0.8573 & 0.8581 & 0.8566 \\
XGBoost & 0.9724 & 0.9098 & 0.9088 & 0.9091 & 0.9087 & 0.9564 & 0.9096 & 0.8486 & 0.8413 & 0.8587 \\
CatBoost & 0.9731 & 0.9088 & 0.9079 & 0.9082 & 0.9076 & 0.9591 & 0.9142 & 0.8552 & 0.8517 & 0.8588  \\
MLP & 0.9638 & 0.8909 & 0.8889 & 0.9122 & 0.8386 & 0.9427 & 0.9009 & 0.8292 & 0.8329 & 0.8256\\
TabTransformer & 0.9649 & 0.8723 & 0.8676 & 0.9477 & 0.7584& 0.9461 & 0.9061 & 0.8360 & 0.8444 & 0.8283  \\
TabKANet & 0.9728 & 0.9102 & 0.9094 & 0.8974 & 0.9042 & 0.9570 & 0.9150 & 0.8577 & 0.8513 & 0.8646 \\

                \bottomrule
	    \end{tabular}}
	\label{table:loosemetstask}
\end{table*}

\subsection*{Numerical Prediction of METS-IR}

Previous studies have confirmed the potential of using the METS-IR index to assess insulin resistance in populations, particularly its impact on diabetes incidence and mortality risk. The calculation of METS-IR requires four indicators: Fasting Plasma Glucose, Triglycerides, BMI, and High-Density Lipoprotein Cholesterol(HDL-C). Our experiments further demonstrated the excellent performance of artificial intelligence (AI) methods in classifying METS-IR based on the thresholds reported in the literature.

Table \ref{table:metsregression} presents the performance of various methods in predicting METS-IR values among nondiabetic individuals. During training, we used the Root Mean Square Error (RMSE) as the evaluation metric. In the internal test using NHANES data, TabKANet achieved the best overall performance, while CatBoost performed slightly worse. In cross-dataset validation using CHARLS data, neural network methods, despite varying complexities, outperformed decision trees and traditional machine learning methods. TabKANet again demonstrated superior performance: in the internal test, it achieved an RMSE of 3.2643 and an R2 of 0.9160; in the external validation, the RMSE was 3.0570 and R2 was 0.8183.

\begin{figure*}[h]
\begin{center}
    \includegraphics[width=\textwidth]{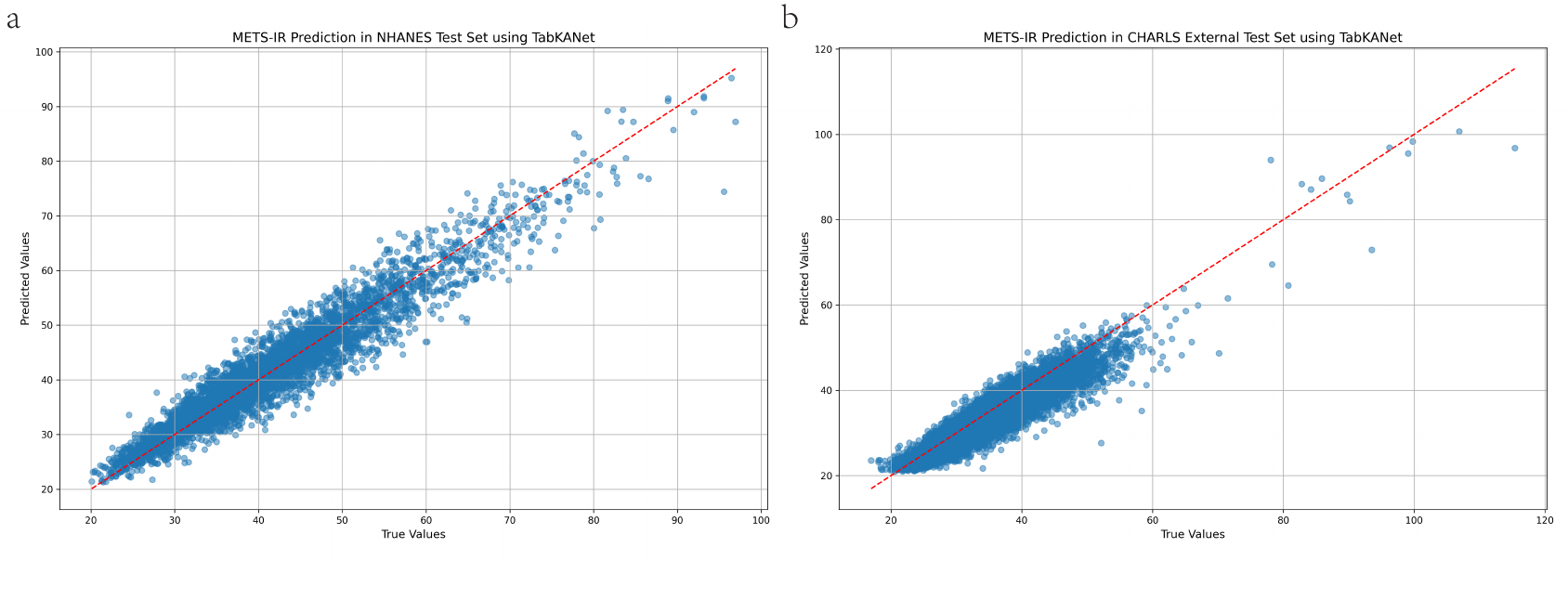} 
\end{center}
\caption{Scatter plot images of METS-IR numerical prediction using TabKANet. a) NHANES test set results. b) CHARLS external test results.}
\label{fig:metsreg}
\end{figure*}

Figure \ref{fig:metsreg} displays the scatter plot of METS-IR values predicted by TabKANet for nondiabetic individuals in the NHANES internal test set and the CHARLS dataset. A significant difference in METS-IR values between the NHANES and CHARLS datasets is evident, which is related to differences in the characteristics of the surveyed populations. For instance, the average BMI in the CHARLS dataset (23.78) is significantly lower than that in the NHANES dataset (28.41). Despite these differences, the model trained on NHANES data can still accurately predict METS-IR values in the CHARLS dataset. The scatter plot clearly demonstrates the high accuracy of the AI method. Additionally, Supplementary Figure 2 in the supplementary file shows the prediction results for different ethnic groups in the NHANES test set, with R2 values ranging from 0.88 to 0.93 across all ethnic groups.




\begin{table*}[h!]
	\footnotesize
	\centering	
    \caption{Comparative Performance in METS-IR Numerical Prediction.}
    \scalebox{1}{
			\begin{tabular}{c|ccc|ccc}
				\toprule
				\multirow{2}{*}{Model}  &\multicolumn{3}{c|}{Internal Test} &\multicolumn{3}{c}{CHARLS(External Test)} \\
				&  MAE & RMSE & R2 &MAE & RMSE & R2 \\
				\midrule  
Linear Regression & 2.5740 & 3.3572 & 0.9026 
 & 4.4707 & 5.2142 & 0.3020\\
Random Forest & 2.5864 & 3.3692 & 0.9022 
& 2.6508 & 3.5640 & 0.6680\\
XGBoost & 2.5348 & 3.3127 & 0.9069 
 & 2.8875 & 3.7950 & 0.6233 \\
CatBoost & 2.5218 & 3.2750 & 0.9091
& 2.6334 & 3.5833 & 0.6619\\
MLP & 2.8538 & 3.7568 & 0.8888 
& 2.3743 & 3.1516 & 0.8069\\
TabTransformer & 2.8139 & 3.7057 & 0.8918 
& 2.3412 & 3.1068 & 0.8123\\
TabKANet & 2.5137 & 3.2643 & 0.9160 
 & 2.2868 & 3.0570 & 0.8183\\ 
                \bottomrule
	    \end{tabular}}
	\label{table:metsregression}
\end{table*}

\subsection*{Impact of features in IR assessment}

Under the METS-IR IR assessment standard, which defines a METS-IR index greater than 41.33 as indicative of potential insulin resistance, the largest number of potential insulin-resistant individuals can be identified from the NHANES database. Among the 22,008 participants in the NHANES database, 9,941 individuals met this criterion for the target population. In the CHARLS dataset, which included 10,333 participants, 1,843 individuals satisfied the condition. Based on this IR assessment standard, we explored the potential impacts of different characteristics under the specified input conditions.

\begin{figure*}[h]
\begin{center}
    \includegraphics[width=\textwidth]{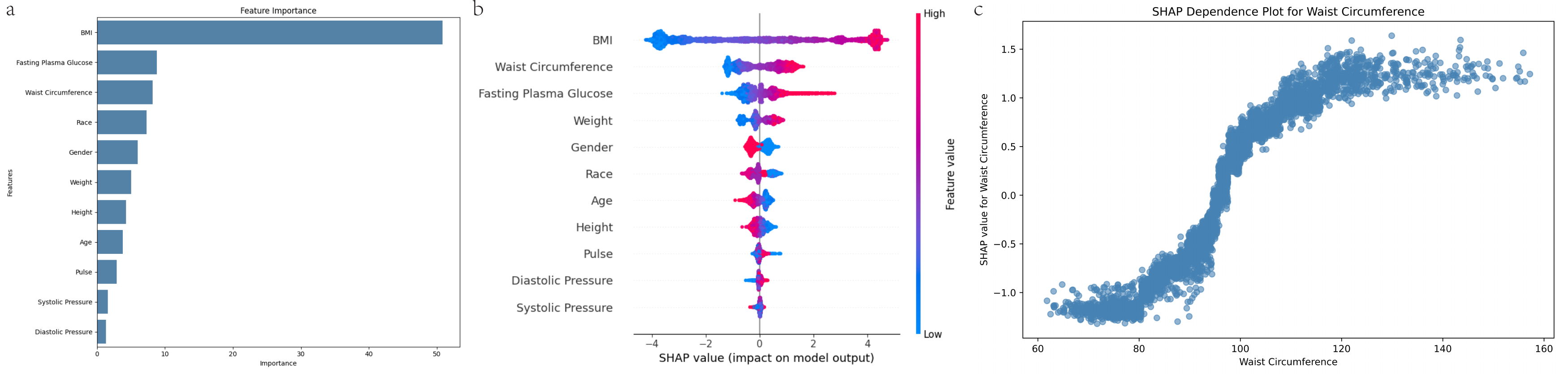} 
\end{center}
\caption{Impact of features in IR assessment based on METS-IR prediction using CatBoost. a) Detailed features importance. b) Impact explanation of features for predicting IR using SHAP values. c) SHAP feature dependency map of waist circumference in IR decision-making.}
\label{fig:impactfea}
\end{figure*}

Figure \ref{fig:impactfea}a illustrates the importance of various features in the assessment of Mets-IR using the CatBoost algorithm. In this specific IR decision, CatBoost achieved an AUC score of 0.9731 and an F1 score of 0.9079. Among the input values, fasting plasma glucose and BMI are already explicitly included in the formula, so it is expected that BMI and fasting plasma glucose would be the top two most important features. However, waist circumference, ethnicity, and gender followed closely behind, even surpassing the influence weight of the age feature.

Figure \ref{fig:impactfea}b shows the impact explanation of features for predicting insulin resistance using SHAP values. Here, BMI and fasting plasma glucose play a crucial role in determining insulin resistance. In addition to these, the role of waist circumference cannot be overlooked. Based on waist circumference, we constructed the SHAP feature dependency map of waist circumference in IR decision-making in Figure \ref{fig:impactfea}c. In this figure, we did not differentiate between genders. For any participant, when the waist circumference exceeds 95 cm, the positive SHAP value of waist circumference in assessing insulin resistance significantly increases. This finding is consistent with the results of several previous studies. A larger waist circumference typically indicates a greater accumulation of visceral fat, which is an important risk factor for insulin resistance and cardiovascular and cerebrovascular diseases\cite{zhao2021relationship,feng2019body}. By measuring individuals' waist circumference, combined with BMI and fasting blood glucose, it is possible to conduct better AI-based assessments of insulin resistance.

\subsection*{Performance of Simplified Models for Resource-Limited Scenarios}




In addition to the research content mentioned above, we have also proposed a simplified prediction model driven by AI. This model uses only BMI and fasting blood glucose as inputs to assess insulin resistance through the best-performing model. In practical clinical practice, doctors often use BMI and fasting blood glucose to empirically determine whether a person has insulin resistance. The experimental results demonstrate that even relying solely on these two indicators, the AI-based approach can still achieve fair model performance. The relevant experimental results are shown in the supplementary document.

In environments with limited resources where waist circumference cannot be measured, the simplified model (BMI + Glucose) shows a performance drop of 1-6 percentage points in AUC across all classification tasks compared to the full model. In terms of METS-IR numerical prediction, this simplified model achieved an internal test RMSE of 3.576 and an external test RMSE of 3.519, which significantly exceeded those of the full model. The superior performance of the full model indicates that it achieves the minimum prediction error with minimal blood testing, which is crucial for guiding personalized interventions.

\section*{Discussion}

In this study, we developed several AI-based models for predicting IR conditions using the NHANES dataset and validated them with the CHARLS dataset. We established stringent data input criteria with fasting blood glucose as the only invasive test required. This approach enabled AI models to perform binary classification or numerical prediction of IR indicators. Our goal is to provide a low-cost, frequently usable, home-based self-testing tool for the broad non-diabetic population, facilitating convenient self-assessment or preliminary screening in medical institutions. We experimentally validated multiple widely used IR assessment criteria and their thresholds and compared the prediction performance of traditional machine learning methods, GBDT methods, and neural network models.

Overall, CatBoost exhibited superior performance in binary IR classification within the general population. Meanwhile, neural network methods, such as TabKANet, demonstrated remarkable capabilities in directly predicting METS-IR values. In internal testing, the best-performing model achieved an R2 of 0.916 and an RMSE of 3.2643. When validated externally using the CHARLS dataset, the model maintained robust performance, with an R2 value of 0.8183 and an RMSE of 3.057.

Our experimental results indicate that in the assessment of IR among the general population, classification models (that is, the binary classification task of determining whether an individual has IR) trained on the NHANES dataset demonstrated good consistency and stability across different models in cross-dataset validation. These experimental results fully demonstrate the great potential of AI methods in assessing IR among the general population. AI can not only build simple binary classification models for IR assessment but also accurately predict specific METS-IR values based solely on fasting blood glucose and basic physical characteristics, thereby providing a more detailed and intelligent solution for IR assessment.

Previous studies have already implemented IR assessments based on feature values\cite{tsai2023development,lee2022development,park2022development}. However, these assessment methods mostly involve invasive blood tests beyond fasting blood glucose, including triglycerides and high-density lipoprotein, among others. Considering the widespread use of glucometers and the convenience of home-based blood glucose monitoring, we believe that to achieve intelligent IR assessment among the general population, the testing items should be streamlined to include only fasting blood glucose as the sole invasive test. For the general population, it would be of greater practical significance if IR could be predicted simply by combining a glucometer with some basic information.

On the other hand, previous research has mostly focused on binary classification assessments of HOMA-IR. Although HOMA-IR is a widely used standard for assessing insulin resistance, recent studies have shown that METS-IR performs better in this regard. The high performance of our proposed solution in predicting METS demonstrates the feasibility of assessing IR based solely on fasting blood glucose. METS-IR takes into account multiple metabolic factors related to IR and can better identify pathological changes associated with IR, especially in predicting visceral fat and type 2 diabetes. In addition, recent studies have also emphasized the ability of METS-IR to predict all-cause mortality and cardiovascular mortality. Therefore, exploring the predictive capabilities of various AI methods across different insulin resistance assessment standards will be more conducive to developing robust IR assessment strategies, thereby achieving broader public health benefits among the general population.

This study explored the importance of features and utilized a METS-IR binary classification model with an AUC value exceeding 0.97. We found that, in addition to BMI and fasting blood glucose, waist circumference also plays an extremely important role in the classification of IR. The SHAP dependence plot of waist circumference, regardless of age and gender, showed that when the waist circumference exceeds 95 cm, its decision importance for the diagnosis of insulin resistance significantly increases. This finding is consistent with the conclusions of multiple studies, indicating that an increase in waist circumference significantly raises the risk of all-cause mortality, as well as the likelihood of developing insulin resistance and diabetes\cite{zhang2024body,chang2015body,feng2019body,liu2021body,zhao2021relationship}.

Our study demonstrates that the AI-driven model for IR assessment performs robustly across diverse racial/ethnic groups, including underrepresented populations such as non-Hispanic Black and Hispanic individuals. This robustness is particularly significant given the higher prevalence of diabetes and cardiovascular disease in these populations, which are often underrepresented in clinical research\cite{2023Social,2009Self}. The model's reliance on universally accessible features, such as fasting blood glucose, BMI, and waist circumference, rather than expensive or region-specific biomarkers (e.g., insulin, advanced lipid profiles), suggests that it could significantly reduce healthcare disparities\cite{Uauy2001Obesity, Francilene2013Normal}. In low- and middle-income countries, where access to laboratory testing is limited and diabetes prevalence is rising rapidly, this simplified tool could serve as a first-line screening method to identify high-risk individuals for early intervention. For example, in sub-Saharan Africa and rural Asia, where laboratory infrastructure is sparse and healthcare budgets are constrained, our model’s requirement for only a glucometer and basic anthropometric measurements aligns with the WHO’s recommendations for scalable non-communicable disease screening\cite{ramani2023prioritising,Uauy2001Obesity, Francilene2013Normal}. By minimizing dependence on complex blood tests and providing low-cost methods for home self-inspection to the maximum extent possible, our approach mitigates the "diagnostic gap" that disproportionately affects socioeconomically disadvantaged populations. Future studies should validate the model in other global populations, such as African and European cohorts, to confirm its universal applicability. In addition, integrating the model into mobile health tools could facilitate widespread adoption, allowing people to monitor their risk of insulin resistance at home and seek timely medical intervention.

In this study, we observed different AUC values when predicting IR using different assessment indicators (HOMA-IR, TyG, and METS-IR) based on the same set of features. This variation likely reflects the distinct physiological mechanisms emphasized by each indicator. The TyG index, calculated from fasting glucose and triglycerides, is closely associated with visceral fat metabolism and may be more sensitive to metabolic abnormalities, such as elevated triglyceride levels, which simple anthropometric features cannot fully capture. In contrast, METS-IR integrates BMI, fasting glucose, triglycerides, and HDL cholesterol, providing a more comprehensive assessment of metabolic syndrome-related features and obesity-associated IR. This holistic approach likely explains its superior AUC value in our study. Our study did not aim to identify a better IR assessment indicator. Instead, our findings highlight the potential of AI-driven methods to identify different IR indicators based on minimally invasive tests. This approach underscores the importance of selecting appropriate IR indicators for specific populations and clinical contexts.

This study has several limitations. First, the lack of insulin measurements in the CHARLS dataset precluded external validation of the HOMA-IR index. Second, CHARLS targets adults aged 45 and above, resulting in an age distribution that differs from the NHANES dataset. Future studies should include external validation across the full adult age range to confirm our findings.

While we selected simple and easily measurable features for IR prediction, other factors, such as lifestyle behaviors (e.g., alcohol consumption and smoking), may also significantly influence IR assessment. These factors are closely related to insulin resistance and metabolic diseases, and incorporating them into future models could further optimize performance.

Additionally, some features used in this study (e.g., blood pressure, pulse, fasting glucose) may be subject to measurement errors or random fluctuations. These errors could affect model performance, especially in resource-limited settings where data quality control is challenging. Future studies should validate the stability of these features under varying conditions and explore data preprocessing or feature engineering techniques to mitigate the impact of errors on model predictions.

Furthermore, although METS-IR demonstrated excellent predictive performance in our study, this advantage needs to be confirmed through clinical validation and longer-term follow-up. Such validation will help assess the stability and reliability of the model’s predictive performance and verify the clinical feasibility and long-term impact of using AI methods for IR assessment.

\section*{Conclusion}

In conclusion, this study aimed to develop an AI-based method for individuals without diabetes to frequently self-assess their insulin resistance. We established stringent data criteria to achieve this, incorporating only fasting blood glucose and easily measurable physical features as input variables. Using multi-ethnic data from NHANES, we developed insulin resistance assessment models based on various AI methods, focusing on the widely used criteria of HOMA-IR, TyG, and METS-IR. The generalizability of these models was further validated using the cross-national CHARLS dataset.

Our results demonstrate that AI methods can effectively predict different insulin resistance criteria using only fasting blood glucose as the invasive test. These models exhibit robust generalizability across datasets. Notably, waist circumference emerged as a significant predictor, playing a crucial role in the prediction outcomes. Importantly, METS-IR stood out with exceptional prediction performance, achieving an AUC of over 0.97. Additionally, neural networks showed strong performance in predicting specific METS-IR values in the non-diabetic population.

These findings provide critical evidence for constructing a low-cost insulin resistance assessment system based on AI methods. Such a system holds great promise for supporting the health management of a broad population, facilitating convenient and accessible self-assessment tools for early detection and prevention of insulin resistance.

\bibliography{aiir}
\bibliographystyle{plain}

\appendix

\section{Appendix}

\subsection{Data set splitting based on HOMA-IR}
\begin{table*}[h]
\centering
\caption*{Supplement Table 1: Characteristics of Data at the HOMA-IR 2.5 Threshold}
\small
\begin{tabular}{lllll}
\hline
Characteristics & NHANES & HOMA-IR$\leq$2.5 & HOMA-IR$>$2.5 & p-value \\
\hline
Number of participants & 22,008 & 12,195 & 9,814 & \\
Age & 46.82$\pm$18.28 & 45.83$\pm$18.41 & 48.03$\pm$18.05 & $<$0.001 \\
Female & 11,433(51.94\%) & 6,465(53.01\%) & 4,968(50.62\%) & $<$0.001 \\
Body mass index(kg/m$^2$) & 28.41$\pm$6.53 & 25.62$\pm$4.71 & 31.87$\pm$6.81 & $<$0.001 \\
Fasting plasma glucose(mg/dl) & 100.64$\pm$18.98 & 95.11$\pm$10.86 & 95.11$\pm$10.86 & $<$0.001 \\
Plasma insulin level(uU/mL) & 12.26$\pm$11.32 & 6.34$\pm$2.33 & 19.60$\pm$13.54 & $<$0.001 \\
Triglycerides(mg/dl) & 117.01$\pm$66.69 & 98.98$\pm$56.17 & 139.4$\pm$71.76 & $<$0.001 \\
HDL Cholesterol(mg/dl) & 54.58$\pm$15.97 & 58.96$\pm$16.52 & 49.13$\pm$13.39 & $<$0.001 \\
Systolic(mm Hg) & 122.44$\pm$18.49 & 120.44$\pm$18.82 & 124.92$\pm$17.76 & $<$0.001 \\
Diastolic(mm Hg) & 70.45$\pm$11.77 & 69.12$\pm$11.37 & 72.09$\pm$12.05 & $<$0.001 \\
Pulse(60 sec.) & 70.66$\pm$11.74 & 69.16$\pm$11.46 & 72.53$\pm$11.82 & $<$0.001 \\
HOMA-IR & 3.17$\pm$3.60 & 1.49$\pm$0.55 & 5.27$\pm$4.56 & $<$0.001 \\
TyG & 8.52$\pm$0.59 & 8.31$\pm$0.53 & 8.77$\pm$0.55 & $<$0.001 \\
METS-IR & 41.68$\pm$11.49 & 36.12$\pm$7.88 & 48.59$\pm$11.53 & $<$0.001 \\
Mexican American & 3,805(17.28\%) & 1,750(14.35\%) & 2,054(20.92\%) & \\
Other Hispanic & 1,868(8.48\%) & 949(7.78\%) & 919(9.36\%) & \\
Non-Hispanic White & 9,688(44.02\%) & 5,800(47.56\%) & 3,888(39.61\%) & \\
Non-Hispanic Black & 4,401(19.99\%) & 2,373(19.46\%) & 2,028(20.66\%) & \\
Other Race-Including Multi-Racial & 2,247(10.21\%) & 1,322(10.84\%) & 925(9.42\%) & \\
\end{tabular}
\end{table*}

\subsection{Data set splitting based on TyG}

\begin{table*}[h]
\centering
\caption*{Supplement Table 2: Characteristics of Data at the TyG 8.85 Threshold}
\small
\begin{tabular}{lllll}
\hline
&Characteristics  & TyG$\leq$8.85 & TyG$>$8.85 & p-value\\
\hline

\multicolumn{5}{l}{\textbf{NHANES}}\\
\hline
Number of participants & 22,008 & 15,660 & 6,348 & \\
Age & 46.82$\pm$18.28 & 45.15$\pm$18.39 & 50.90$\pm$17.36 & $<$0.001 \\
Female & 11,433(51.94\%) & 8,515(54.37\%) & 2,918(45.96\%) & $<$0.001 \\
Body mass index(kg/m$^2$) & 28.41$\pm$6.53 & 27.58$\pm$6.48 & 30.44$\pm$6.19 & $<$0.001 \\
Fasting plasma glucose(mg/dl) & 100.64$\pm$18.98 & 97.20$\pm$11.13 & 109.13$\pm$29.02 & $<$0.001 \\
Plasma insulin level(uU/mL) & 12.26$\pm$11.32 & 10.31$\pm$8.76 & 17.06$\pm$14.91 & $<$0.001 \\
Triglycerides(mg/dl) & 117.01$\pm$66.69 & 83.51$\pm$29.69 & 199.63$\pm$60.43 & $<$0.001 \\
HDL Cholesterol(mg/dl) & 54.58$\pm$15.97 & 57.92$\pm$15.97 & 46.33$\pm$12.63 & $<$0.001 \\
Systolic(mm Hg) & 122.44$\pm$18.49 & 120.84$\pm$18.25 & 126.37$\pm$18.50 & $<$0.001 \\
Diastolic(mm Hg) & 70.45$\pm$11.77 & 69.73$\pm$11.53 & 72.21$\pm$12.17 & $<$0.001 \\
Pulse(60 sec.) & 70.66$\pm$11.74 & 70.03$\pm$11.49 & 72.22$\pm$12.19 & $<$0.001 \\
HOMA-IR & 3.17$\pm$3.60 & 2.54$\pm$2.58 & 4.74$\pm$5.01 & $<$0.001 \\
TyG & 8.52$\pm$0.59 & 8.23$\pm$0.40 & 9.23$\pm$0.30 & $<$0.001 \\
METS-IR & 41.68$\pm$11.49 & 38.89$\pm$10.45 & 48.56$\pm$11.04 & $<$0.001 \\
Mexican American & 3,805(17.28\%) & 2,380(15.19\%) & 1,424(22.43\%) & \\
Other Hispanic & 1,868(8.48\%) & 1,259(8.03\%) & 609(9.59\%) & \\
Non-Hispanic White & 9,688(44.02\%) & 6,640(42.40\%) & 3,048(48.01\%) & \\
Non-Hispanic Black & 4,401(19.99\%) & 3,771(24.08\%) & 630(9.92\%) & \\
Other Race - Including Multi-Racial & 2,247(10.21\%) & 1,610(10.28\%) & 637(10.03\%) & \\
\hline
\multicolumn{5}{l}{\textbf{CHARLS}}\\
\hline
Number of participants & 1,0333 & 7,128 & 3,205 & \\
Age & 60.20$\pm$9.87 & 60.41$\pm$10.08 & 59.73$\pm$9.37 & $<$0.001 \\
Female & 5,544(53.65\%) & 3,712(52.07\%) & 1,832(57.16\%) & $<$0.001 \\
Body mass index(kg/m$^2$) & 23.78$\pm$3.84 & 23.13$\pm$3.73 & 25.23$\pm$3.69 & $<$0.001 \\
Fasting plasma glucose(mg/dl) & 93.97$\pm$11.76 & 91.27$\pm$11.01 & 99.98$\pm$11.13 & $<$0.001 \\
Plasma insulin level(uU/mL) & - & - & - & $<$0.001 \\
Triglycerides(mg/dl) & 136.27$\pm$85.59 & 93.51$\pm$27.92 & 231.37$\pm$93.68 & $<$0.001 \\
HDL Cholesterol(mg/dl) & 51.66$\pm$11.37 & 53.85$\pm$11.56 & 46.79$\pm$9.24 & $<$0.001 \\
Systolic(mm Hg) & 127.27$\pm$19.33 & 125.90$\pm$19.33 & 130.32$\pm$18.98 & $<$0.001 \\
Diastolic(mm Hg) & 75.35$\pm$11.22 & 74.33$\pm$11.08 & 77.62$\pm$11.19 & $<$0.001 \\
Pulse(60 sec.) & 73.58$\pm$10.45 & 72.83$\pm$10.34 & 75.27$\pm$10.52 & $<$0.001 \\
HOMA-IR & - & - & - & $<$0.001 \\
TyG & 8.60$\pm$0.56 & 8.30$\pm$0.32 & 9.28$\pm$0.35 & $<$0.001 \\
METS-IR & 35.14$\pm$7.17 & 32.92$\pm$6.11 & 40.05$\pm$6.90 & $<$0.001 \\

\end{tabular}
\end{table*}

\subsection{Data set splitting based on METS-IR}
\begin{table*}[h!]
\centering
\caption*{Supplement Table 3: Characteristics of Data at the METS-IR 41.33 Threshold}
\small
\begin{tabular}{lllll}
\hline
&Characteristics  & METS-IR$\leq$41.33 & METS-IR$>$41.33 & p-value\\
\hline
\multicolumn{5}{l}{\textbf{NHANES}}\\
\hline
Number of participants & 22,008 & 12,067 & 9,941 & \\
Age & 46.82$\pm$18.28 & 46.15$\pm$19.20 & 47.61$\pm$17.07 & $<$0.001 \\
Female & 11,433(51.94\%) & 6,552(54.29\%) & 4,881(49.09\%) & $<$0.001 \\
Body mass index(kg/m$^2$) & 28.41$\pm$6.53 & 24.13$\pm$3.07 & 33.59$\pm$5.82 & $<$0.001 \\
Fasting plasma glucose(mg/dl) & 100.64$\pm$18.98 & 96.61$\pm$12.78 & 105.53$\pm$23.58 & $<$0.001 \\
Plasma insulin level(uU/mL) & 12.26$\pm$11.32 & 8.11$\pm$6.21 & 17.30$\pm$13.80 & $<$0.001 \\
Triglycerides(mg/dl) & 117.01$\pm$66.69 & 94.53$\pm$50.00 & 144.29$\pm$73.85 & $<$0.001 \\
HDL Cholesterol(mg/dl) & 54.58$\pm$15.97 & 61.14$\pm$16.16 & 46.62$\pm$11.49 & $<$0.001 \\
Systolic(mm Hg) & 122.44$\pm$18.49 & 120.50$\pm$19.05 & 124.79$\pm$17.52 & $<$0.001 \\
Diastolic(mm Hg) & 70.45$\pm$11.77 & 68.74$\pm$11.39 & 72.52$\pm$11.90 & $<$0.001 \\
Pulse(60 sec.) & 70.66$\pm$11.74 & 69.90$\pm$11.65 & 71.60$\pm$11.77 & $<$0.001 \\
HOMA-IR & 3.17$\pm$3.60 & 1.97$\pm$1.74 & 4.63$\pm$4.60 & $<$0.001 \\
TyG & 8.52$\pm$0.59 & 8.29$\pm$0.51 & 8.79$\pm$0.55 & $<$0.001 \\
METS-IR & 41.68$\pm$11.49 & 33.53$\pm$4.88 & 51.58$\pm$9.21 & $<$0.001 \\
Mexican American & 3,805(17.28\%) & 1,784(14.78\%) & 2,020(20.31\%) & \\
Other Hispanic & 1,868(8.48\%) & 920(7.62\%) & 948(9.53\%) & \\
Non-Hispanic White & 9,688(44.02\%) & 5,484(45.44\%) & 4,204(42.28\%) & \\
Non-Hispanic Black & 4,401(19.99\%) & 2,326(19.27\%) & 2,075(20.87\%) & \\
Other Race-Including Multi-Racial & 2,247(10.21\%) & 1,553(12.86\%) & 694(6.98\%) & \\
\hline
\multicolumn{5}{l}{\textbf{CHARLS}}\\
\hline
Number of participants & 10,333 & 8,490 & 1,843 & \\
Age & 60.20$\pm$9.87 & 60.60$\pm$9.97 & 58.35$\pm$9.17 & $<$0.001 \\
Female & 5,544(53.65\%) & 4,491(52.89\%) & 1,053(57.13\%) & $<$0.001 \\
Body mass index(kg/m$^2$) & 23.78$\pm$3.84 & 22.65$\pm$2.74 & 28.98$\pm$3.90 & $<$0.001 \\
Fasting plasma glucose(mg/dl) & 93.97$\pm$11.76 & 92.93$\pm$11.53 & 98.76$\pm$11.61 & $<$0.001 \\
Plasma insulin level(uU/mL) & - & - & - & $<$0.001 \\
Triglycerides(mg/dl) & 136.27$\pm$85.59 & 118.66$\pm$65.32 & 217.38$\pm$115.83 & $<$0.001 \\
HDL Cholesterol(mg/dl) & 51.66$\pm$11.37 & 53.47$\pm$11.19 & 43.31$\pm$7.94 & $<$0.001 \\
Systolic(mm Hg) & 127.27$\pm$19.33 & 126.13$\pm$19.28 & 132.51$\pm$18.69 & $<$0.001 \\
Diastolic(mm Hg) & 75.35$\pm$11.22 & 74.47$\pm$11.02 & 79.40$\pm$11.24 & $<$0.001 \\
Pulse(60 sec.) & 73.58$\pm$10.45 & 73.42$\pm$10.48 & 74.32$\pm$10.29 & $<$0.001 \\
HOMA-IR & - & - & - & $<$0.001 \\
TyG & 8.60$\pm$0.56 & 8.49$\pm$0.49 & 9.14$\pm$0.54 & $<$0.001 \\
METS-IR & 35.14$\pm$7.17 & 32.72$\pm$4.77 & 46.24$\pm$5.78 & $<$0.001 \\

\end{tabular}
\end{table*}

\subsection{Differences in IR prediction among different races}

\begin{figure*}[h!]
\begin{center}
    \includegraphics[width=\textwidth]{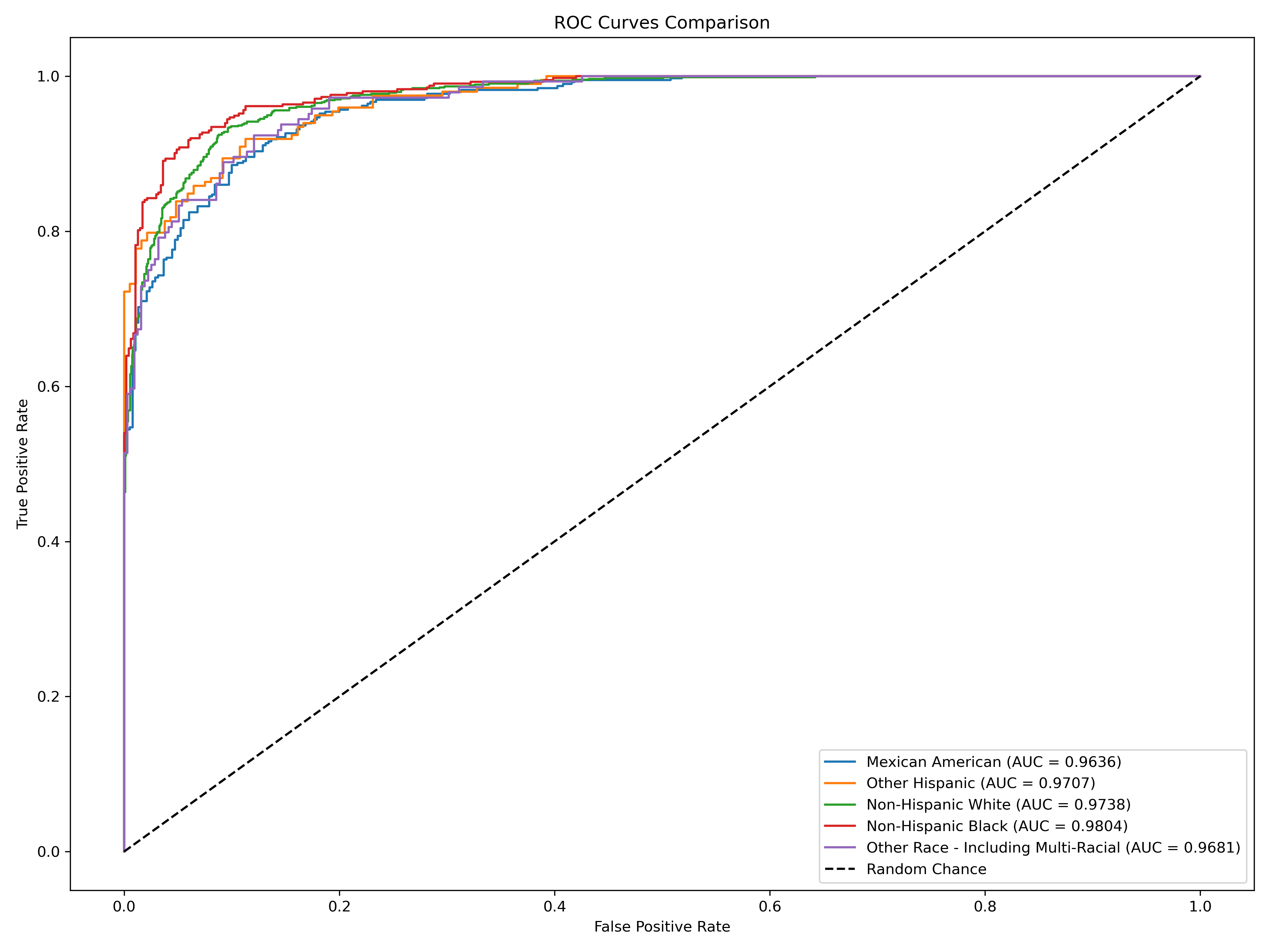} 
\end{center}

\caption*{Supplement Figure 1: ROC curves for METS-IR classification by race/ethnicity using CatBoost. The model demonstrated high performance across all subgroups, with AUC values ranging from 0.96 to 0.97.}
\end{figure*}

\begin{table*}[h!]
\centering

\caption*{Supplement Table 4: Performance Metrics for Different Races/Ethnicities}

\label{tab:performance_metrics}
\begin{tabular}{lccc}
\toprule
\textbf{Race/Ethnicity} & \textbf{MAE} & \textbf{RMSE} & \textbf{R$^2$} \\
\midrule
Mexican American        & 2.716 & 3.507 & 0.887 \\
Other Hispanic         & 2.553 & 3.157 & 0.914 \\
Non-Hispanic White     & 2.553 & 3.300 & 0.917 \\
Non-Hispanic Black     & 2.410 & 3.163 & 0.932 \\
Other Race - Including Multi-Racial & 2.174 & 2.953 & 0.904 \\
\bottomrule
\end{tabular}
\end{table*}

\begin{figure*}[h!]
\begin{center}
    \includegraphics[width=\textwidth]{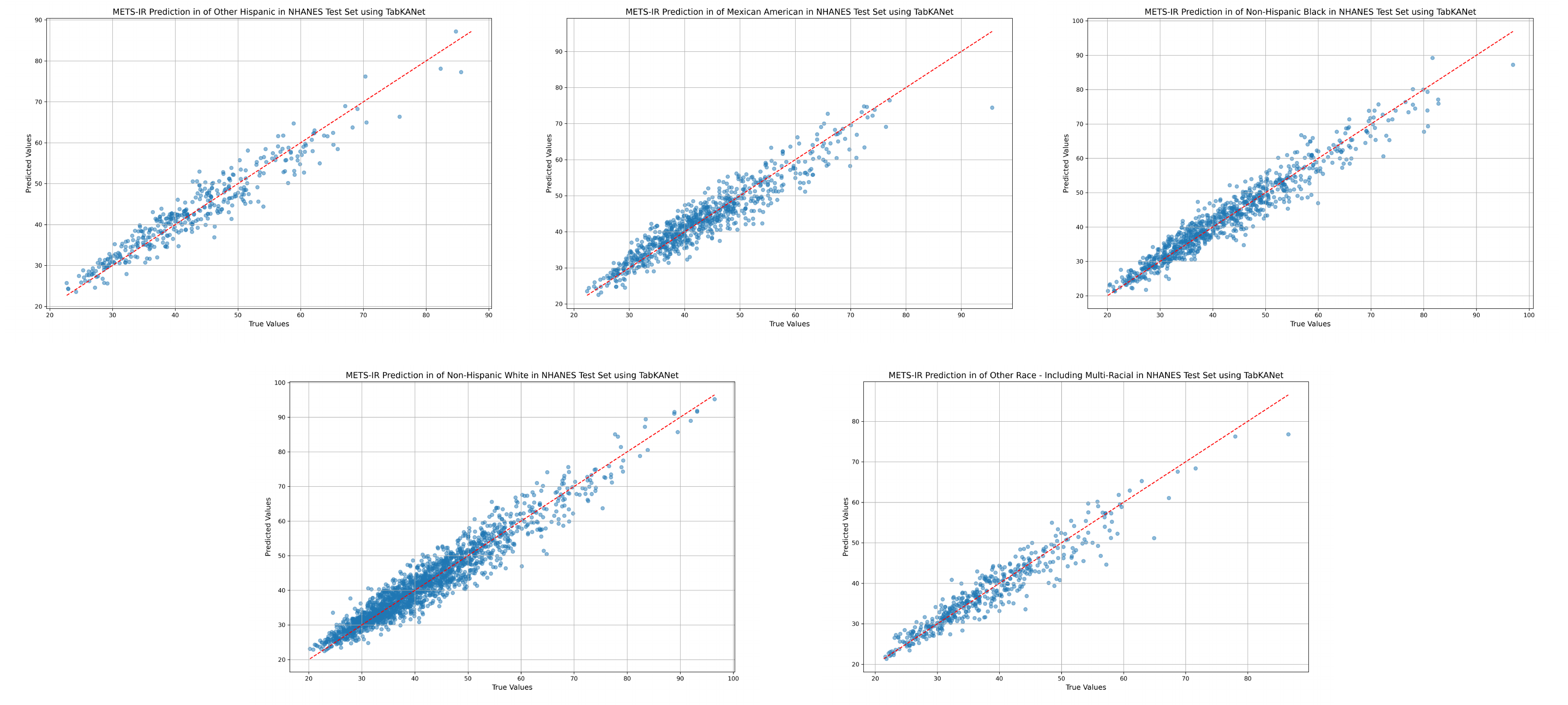} 
\end{center}

\caption*{Supplement Figure 2: Scatter plots of predicted vs. actual METS-IR values by race/ethnicity using TabKANet. The model achieved high accuracy across all subgroups, with R$^2$ values ranging from 0.88 to 0.93.}
\end{figure*}

\subsection{Comparison of performance between the Simplified Model and the Full Model}

In this article, we define BMI and fasting blood glucose conditions as the Simplified Model and compare the performance of the Simplified Model with our proposed Full Model.

\begin{table*}[h!]
	\footnotesize
	\centering	

\caption*{Supplement Table 5: The performance of the Simplified Model in HOMA-IR task.}

	\scalebox{1}{
			\begin{tabular}{c|c c c c c}
				\toprule
		Model & AUC & ACC & F1 & Precision & Recall \\
				\midrule  
Simplified (BMI+Glucose)  & 0.8321 & 0.7643 & 0.7604 & 0.7622 & 0.7594 \\
Full Model & 0.8595 & 0.7802 & 0.7773 & 0.7779 & 0.7768 \\
                \bottomrule
	    \end{tabular}}
\end{table*}

\begin{table*}[h!]
	\footnotesize
	\centering	

\caption*{Supplement Table 6: The performance of the Simplified Model in TyG task.}

    \scalebox{1}{
			\begin{tabular}{c|ccccc|ccccc}
				\toprule
                
				\multirow{2}{*}{Model}  &\multicolumn{5}{c|}{Internal Test} &\multicolumn{5}{c}{CHARLS(External Test)} \\
				& AUC & ACC & F1 & Precision & Recall&AUC & ACC & F1 & Precision & Recall \\
				\midrule  
Simplified (BMI+Glucose)   & 0.7358 & 0.7402 & 0.4592 & 0.7349 & 0.5231 & 0.7143 & 0.7400 & 0.5963 & 0.6891 & 0.5938 \\
Full Model &  0.7777 & 0.7564 & 0.6568 & 0.7056 & 0.6437 &  0.7442 & 0.7064 & 0.6603 & 0.6589 & 0.6620\\
                \bottomrule
	    \end{tabular}}
\end{table*}

\begin{table*}[h!]
	\footnotesize
	\centering	

\caption*{Supplement Table 7: The performance of the Simplified Model in METS-IR task.}

    \scalebox{1}{
			\begin{tabular}{c|ccccc|ccccc}
				\toprule
                
				\multirow{2}{*}{Model}  &\multicolumn{5}{c|}{Internal Test} &\multicolumn{5}{c}{CHARLS(External Test)} \\
				& AUC & ACC & F1 & Precision & Recall&AUC & ACC & F1 & Precision & Recall \\
				\midrule  
Simplified (BMI+Glucose)  &  0.9590 & 0.9023 & 0.8575 & 0.8438 & 0.8736 & 0.9480 & 0.9031 & 0.8388 & 0.8312 & 0.8465 \\
Full Model & 0.9731 & 0.9088 & 0.9079 & 0.9082 & 0.9076 & 0.9591 & 0.9142 & 0.8552 & 0.8517 & 0.8588\\
                \bottomrule
	    \end{tabular}}
\end{table*}

\begin{table*}[h!]
	\footnotesize
	\centering	
\caption*{Supplement Table 8: The performance of the Simplified Model in METS-IR numerical prediction.}

    \scalebox{1}{
			\begin{tabular}{c|ccc|ccc}
				\toprule
				\multirow{2}{*}{Model}  &\multicolumn{3}{c|}{Internal Test} &\multicolumn{3}{c}{CHARLS(External Test)} \\
				&  MAE & RMSE & R2 &MAE & RMSE & R2 \\
				\midrule  
Simplified (BMI+Glucose) & 2.7463 & 3.5760 & 0.8992 & 2.8050 & 3.5190 & 0.8634 \\
Full Model & 2.5137 & 3.2643 & 0.9160 & 2.2868 & 3.0570 & 0.8183 \\
                \bottomrule
	    \end{tabular}}
\end{table*}

\end{document}